\pdfoutput=1
\documentclass[11pt]{article}

\usepackage[]{ACL2023}

\usepackage{times}
\usepackage{latexsym}

\usepackage[T1]{fontenc}
\usepackage[utf8]{inputenc}

\usepackage{microtype}

\usepackage{inconsolata}

\usepackage{microtype}
\usepackage{booktabs}
\usepackage{makecell}
\usepackage{graphicx}
\usepackage{mathtools}
\usepackage{enumitem}
\usepackage{float}
\usepackage{colortbl}
\usepackage[most]{tcolorbox}
\usepackage{pifont}
\usepackage{multirow}
\usepackage{amsmath}
\usepackage{amssymb}
\usepackage{subcaption}
\usepackage{hyperref}
\usepackage[toc,page]{appendix}
\usepackage{calc}
\usepackage{scrextend}
\usepackage{bbm}

\hypersetup{colorlinks,allcolors=black}
\DeclareMathOperator*{\argmax}{argmax}  
\newcommand{\RNum}[1]{\uppercase\expandafter{\romannumeral #1\relax}}

\newcommand{\ie}{\textit{i}.\textit{e}.}
\newcommand{\eg}{\textit{e}.\textit{g}.}
\newcommand{\kyungjae}[1]{\textcolor{blue}{#1}}
\newcommand{\ourmodel}{UR}

\title{When to Read Documents or QA History: 
\\ On Unified and Selective Open-domain QA}

\author{Kyungjae Lee$^{1}$\thanks{ ~~First two authors equally contributed to this work.} \quad\quad Sang-eun Han$^{2,3*}$  \quad\quad Seung-won Hwang$^{2,3}$\thanks{~~correspond to seungwonh@snu.ac.kr}  \quad\quad Moontae Lee$^{1,4}$  \\  \quad $^1$LG AI Research \quad\quad $^2$SNU-LG AI Research Center
 \\ $^3$Seoul National University \quad\quad $^4$University of Illinois at Chicago}

\begin{document}
\maketitle

\begin{abstract}
This paper studies the problem of open-domain question answering, with the aim of answering a diverse range of questions leveraging knowledge resources. Two types of sources, QA-pair and document corpora, have been actively leveraged with the following complementary strength. The former is highly precise when the paraphrase of given question $q$ was seen and answered during training, often posed as a retrieval problem, while the latter generalizes better for unseen questions. A natural follow-up is thus leveraging both models, while a naive pipelining or integration approaches have failed to bring additional gains over either model alone. Our distinction is interpreting the problem as calibration, which estimates the confidence of predicted answers as an indicator to decide when to use a document or QA-pair corpus. The effectiveness of our method was validated on widely adopted benchmarks such as Natural Questions and TriviaQA.
%and we will publicly release our code and settings.
\end{abstract}

\section{Introduction}

Open-domain question answering is a well-known task in natural language processing, aiming to answer factoid questions from an open set of domains.
One commonly used approach for this task is the retrieve-then-read pipeline (also known as \textit{Open-book QA}) to retrieve relevant knowledge, then reason answers over the knowledge.
Given the wide range of topics that open-domain questions can cover, a key to a successful answering model is: to access and utilize diverse knowledge sources effectively.

% “any external information absent from the input, but helpful for generating the output”.

Toward this goal, existing work can be categorized by the knowledge source used:
\begin{itemize}[leftmargin=0.4cm]
\item Document Corpus-based QA (\textbf{Doc-QA}): This type of work utilizes
a general-domain \textbf{Document Corpus} (\eg, Wikipedia)~\cite{karpukhin2020dense,guu2020retrieval,liu2021dense,izacard2021leveraging} 
 for reading then answering questions (\ie, $\{Q,D\}\rightarrow A$).
 \item QA as Retrieval (\textbf{QR}):
 This type of work utilizes a collection of already answered questions (or QA-pair) as knowledge, typically
 leveraging nonparametric approaches, such as a
 retriever for closest QA-pairs, to extract the top-1 QA pair that is most similar to a target question and is considered as a final answer \cite{lewis2021paq,xiao2021open,lewis2021question}. \ie, $Q\rightarrow \{$paraphrase $Q', A\}$.

\end{itemize}

% corresponding to target question as a final answer.

% match a target question with paraphrased one (with an answer) in the QA corpus (\ie, $Q\rightarrow \{$paraphrase $Q', A\}$).
% \textit{Question Retrieval} (Q-Retrieval) \cite{lewis2021paq}:  matching a given question with paraphrased one (with an answer) in existing QA corpus (\ie, $Q\rightarrow \{$paraphrase $Q', A\}$).
% Beside, some works used tabular data \cite{} and knowledge graph~\cite{} for open-domain QA.

% However, as in \cite{lewis2021question}, CB model relies heavily on memorization, which is able to answer questions that overlaps (or paraphrase) with training set, while failing on no-overlap examples.
% That is, CB setting is suitable to match target question $Q$ with a paraphrased $Q'$ in training set, not generalization on unseen questions.

\begin{figure*}[t]
	\centering
	\includegraphics[width=158mm]{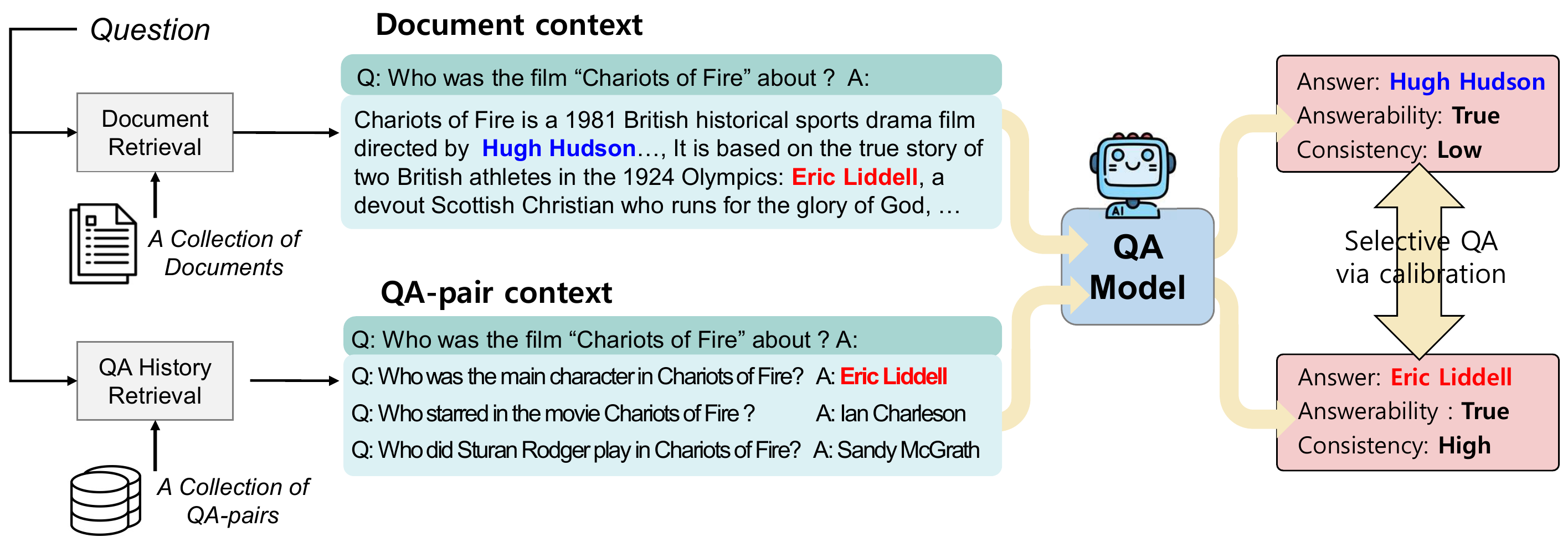}
	\caption{An overview of our Unified Reader QA. We retrieve contexts from document and QA-pair corpus, infer answers from each source, then select the final answer by comparing the calibrated confidences.}
	\label{figure1}
\end{figure*}

In an effort to leverage complementary strengths of existing models, previous work has attempted to build a pipeline of individual models \cite{lewis2021paq}. However, their approach has not resulted in significant gains over using either model alone. 
In this paper, we propose a novel approach of leveraging the strengths of both document and QA pairs as contexts for a \textbf{Unified Reader}-based QA (or \textbf{UR-QA}).\footnote{We stress that our focus is a unified framework, and orthogonal to optimizing readers or retrievers, which is beyond the scope of this paper.}
Figure \ref{figure1} illustrates the distinction of our approach  providing both knowledge to a unified reader as context.
We retrieve a list of relevant QA-pairs (called as \textbf{QA-history}), then treat the few retrieved QA examples, as if it is a relevant document passage.

% Figure \ref{figure1} illustrates the distinction of our approach, treating the few retrieved QA examples, as if it is relevant document passages and providing them to a unified reader as context.
Meanwhile, the closest approach to use multiple knowledge sources is concatenating the multi-sources uniformly into a single decoder \cite{oguz2020unik}, but we argue \textbf{knowledge selection} is critically missing.
To motivate, Figure \ref{figure1} shows the QA-history, from which answer `Eric Liddell' is explicitly identified, 
while it is more implicit in the document such that another name such as `Hugh Hudson' is known to often confuse QA models.
It is critical for the QA model to \textbf{calibrate} prediction quality as an indicator to decide when to use a document corpus or QA-history.

Toward the goal, we propose Selective QA, where a more reliable answer among candidates can be identified through the calibration of the QA model.
Existing calibration~\cite{kamath2020selective,zhang2021knowing,si2022revisiting} %aims to %calibrate the model's outputs (or ``knows when it doesn’t know''), so that QA model can abstain from answering if the model doesn't know.
has focused on the ability of models to ``know when they don't know'' and abstain from answering if they are uncertain. A naive approach would be simply prioritizing more confident predictions for answer selection.

As a known measure of confidence, LM likelihood of generated tokens has been found to often miscalibrate~\cite{jiang2021can,kumar2019calibration}, tending to prefer short outputs \cite{murray2018correcting}, or being biased towards more frequent words \cite{ott2018scaling}.
We also observed similar issues in our setting, which we refer to as \textbf{calibration overfitting} -- LM likelihoods are biased towards increasing confidence on both correct and wrong answers.

Our distinction is to overcome this limitation, by proposing two new objectives, for lowering confidence when the given context cannot answer the question (\ie, answerability), or when sampling uncertainty from decoder is high (\ie, sampling consistency).
%proposing (b) Answerability and (c) Sampling Consistency as confidence calibration.
%We argue the reliability of predicted answers also relies on Knowledge uncertainty (a given knowledge may not contain an answer) and Decoding uncertainty (unselected tokens from decoder may have high confidence).
%To model these uncertainties, we train Answerability and Sampling Consistency as linguistic expressions, jointly with QA training.
% incorrect generic answers such as “I don’t know” are often assigned high probability, 
% relegating the desired answers to the tail of the distribution where softmax is poorly calibrated
Finally, building upon improved calibration, we carefully select among answer candidates inferred from document and QA-pairs.

% Analysis will be Added  - Two sources have been studied to be complementary—The former is highly precise when the given question q was seen during training (in its paraphrased form q′) and answered, while the latter is more effective for unseen q. 

To summarize, we make the following contributions:
a) We propose an open-domain QA model complementing document corpus with QA-pair corpus, and decide the selective usage between a document or QA-pair corpus through calibration.
% and select a final answer among candidates obtained from each knowledge (Section 3).
b) We evaluate our approach on Natural Questions~\cite{kwiatkowski2019natural} and TriviaQA \cite{joshi2017triviaqa}, and our method can improve QA performance of existing models.
c) We analyze how our method improves calibration and how it helps to select better answers.

\section{Related Works}

\textbf{Doc-QA} has been a dominant paradigm in open-domain QA \cite{karpukhin2020dense,guu2020retrieval,liu2021dense,izacard2021leveraging}, where the relevant passages are first fetched by the \emph{retriever} model and then processed by the \emph{reader} model to produce the answer.
\emph{Reader} models are typically categorized as an \emph{extractive} or \emph{generative} model, where the former locates the answer span in the given context and the latter generates the answer in token-by-token manner. 
In our work, we focus on a \emph{generative} model, which can transfer knowledge from generative LMs such as T5 \cite{raffel2020exploring} and GPT-3 \cite{brown2020language}.
% Some recent works have proposed methods to integrate both models \cite{cheng2021unitedqa,fajcik2021r2} in order to leverage the complementarity between two inference methods.
Meanwhile, while most works for open-domain QA use Wikipedia as context, some works~\cite{oguz2020unik,ma2022open} leverage various knowledge including Tables and Knowledge Graphs.

% but we approach in knowledge perspective. 
% In this work, we use generative model based on T5 \cite{raffel2020exploring} and FiD \cite{izacard2021leveraging} to handle both knowledge matching (QA pairs) and reading comprehension (text passages). 
% We infer candidate answers upon each type of data and then select the final answer from candidates, rather than concatenating all types of data into single input as in previous works on knowledge integration \cite{oguz2020unik,ma2022open}.

\textbf{QR} retrieving relevant QA pairs over a large collection of QA pairs is a more efficient alternative to Doc-QA.
\citet{lewis2021paq} build PAQ (for Probably Asked Questions) -- 65M QA pairs: automatically-generated resources by using question generation techniques, and learn RePAQ retriever to efficiently extract the top-1 QA pair that is most similar to a target question, and uses its answer for answering the question.
\citet{xiao2021open} use answer aggregation heuristic to combine retrieved candidates of QA-pairs with candidates from other sources.
\citet{chen2022augmenting} also leverage retrieved QA-pairs, by fusing representations of the QA-pairs into language models.
Despite some gains, their generalizability for unseen questions is limited, compared to Doc-QA, which motivates our approach of selectively combining with other knowledge. 

% REINA \cite{wang2022training} retrieves the closest examples from training set and concatenates them to the input question for language model, which improves the performance on QA tasks such as CommonsenseQA  \cite{talmor2019commonsenseqa}. 
% QR-QA views QA pairs as the source of knowledge for question answering, in such a way that text passages, tables and knowledge graphs are.

\textbf{Our Distinction} is to analyze and utilize the complementarity of Doc-QA and QR, carefully selecting knowledge sources via calibration, while the previous work \cite{oguz2020unik} blindly concatenates all types of data into a single context.

Calibration has been studied for abstaining from answering when the model does not know.
Sources of calibration have been LM's likelihoods \cite{si2022revisiting}, classifier \cite{kamath2020selective}, and linguistic expressions \cite{lin2022teaching,mielke2022reducing,kadavath2022language,tian2023strategies}.
Our distinction is exploring the use of calibration for selective QA, and overcoming the calibration overfitting we observed from existing methods, by proposing new likelihoods based on answerability and consistency.

\section{Proposed Method}
In this section, we formally describe Doc-QA as backbones (Section 3.1) and our unification baseline (Section 3.2), followed by our proposed calibration for selective QA (Section 3.3).

\subsection{Backbone: Doc-QA}

Open-book QA requires to answer question $q$ given context $\mathbf{c}$, \ie, optimizing $\text{P}_{LM}(\mathbf{a}|\mathbf{q}, \mathbf{c})$.
Doc-QA~\cite{lee2019latent,karpukhin2020dense} typically uses Wikipedia documents as knowledge $\mathbf{c}$.
%In practice, 
%\citet{karpukhin2020dense} leverage dense retrieval to retrieve top-$n$ passages $\mathbf{d}_{1:n}$ corresponding to $\mathbf{q}$ and use them as context.

In this paper, for implementing a Doc-QA backbone, we use a state-of-the-art generative reader: Fusion-in-Decoder \cite{izacard2021leveraging}, based on a pretrained language model -- T5~\cite{raffel2020exploring}.
This approach separately encodes top-$n$ passages in an encoder, and fuses them in a decoder.
The final answer $\text{A}$ is obtained as follows:
\begin{equation}
    \begin{split}
    \text{Fuse}(\mathbf{q}, \mathbf{d}_{1:n}) = & ~ [\text{Enc}(\mathbf{q}, \mathbf{d}_{1});,...,;\text{Enc}(\mathbf{q}, \mathbf{d}_{n})] \\
    \text{A} = & ~ \text{Dec (\text{Fuse}}(\mathbf{q}, \mathbf{d}_{1:n}))
    \end{split}
\end{equation}
% \vspace{-2mm}
where Enc and Dec indicate Encoder and Decoder modules in transformer \cite{vaswani2017attention}, and $[~; ]$ indicates the concatenation of encoder's outputs.
Let $\mathbf{x}$ denote the input sequence, and $\mathbf{y} = (y_1,...,y_T)$ the output sequence.
The language model based QA model is trained with maximum likelihood estimation (MLE) to optimize the following objective for a given ($\mathbf{x}$, $\mathbf{y}$): 
\begin{equation}
    \begin{split}
    \mathcal{L}(\mathbf{x},\mathbf{y}) = - \sum_{t=1}^{T} \text{log}~\text{P}_{LM}( y_t |\mathbf{y}_{<t}, \mathbf{x})
    \end{split}
    \label{lm_loss}
\end{equation}
where $\mathbf{x}$ is a pair of question/document $(\mathbf{q},\mathbf{d}_{1:n})$, and $\mathbf{y}$ is the ground-truth answer $\mathbf{a}^*$ in our setting.
Meanwhile, at inference time, we use Greedy Decoding,\footnote{As a decoding method, we can choose beam search or temperature-based sampling, but greedy decoding empirically outperformed others in our QA tasks.} which is commonly used for QA tasks.
A decoded sequence is $\mathbf{\hat{a}}=(\hat{a}_1,\hat{a}_2,...,\hat{a}_T)$, where each token is selected as follows:
\begin{equation}
    \begin{split}
    \hat{a}_t = \argmax_{y \in V} ~ \text{P}_{LM}(y|\mathbf{\hat{a}}_{<t}, \mathbf{q}, \mathbf{d}_{1:n})
    \end{split}
    \label{eq3}
\end{equation}

\begin{figure}[t!]
	\centering
	\includegraphics[width=74mm]{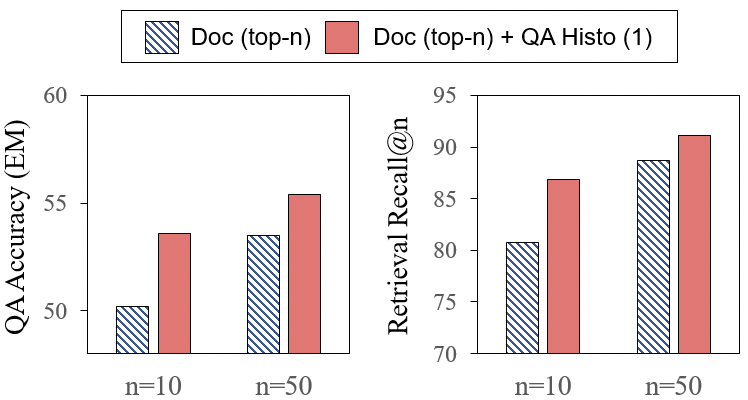}
    % \vspace{-3mm}
	\caption{The results of QA (Left) and Retrieval (Right) on NQ.}
	\label{figure_preliminary}
\end{figure}

\subsection{Unified Reader: UR-QA}

While traditional methods rely on high-efficiency retrievers to match questions with QA history,
our work is
inspired by \textit{in-context learning} \cite{brown2020language} for closed-book QA: 
We propose using the QA-history retrieved as a hypothetical document with few-shot examples and reading it to answer the question

As shown in Figure \ref{figure1}, we retrieve top-$n$ QA pairs from QA corpus as in-context examples, 
and finetune a QA model with the in-context examples.
Specifically, as QA corpus and QR, we used PAQ and a dense retrieval of RePAQ (See Experimental Section for more details), as proposed in \citet{lewis2021paq}.
Given a target question, we extract top-$m$ QA-pairs from PAQ and 
the top-$m$ retrieved QA-pairs, as they are short, can be concatenated into one document passage
as below:
%are provided as context, as if it is a passage retrieved from Doc-QA (denoted as $\mathbf{k}$) as below:
\begin{addmargin}[0.6em]{0em}
{\small \texttt{\noindent Question: \kyungjae{\{target $q$\}}, ~ Answer: ~~~~~~~~~~~~ $\backslash$n \\  Question: \kyungjae{\{example $q_1$\}}, Answer: \kyungjae{\{example $a_1$\}} $\backslash$n \\  Question: \kyungjae{\{example $q_2$\}}, Answer: \kyungjae{\{example $a_2$\}} $\backslash$n \\ Question: ~~~ ... ~~~~~ Answer: ~~~ ...}}
\end{addmargin}
% \vspace{2mm}

% To unify the two backbones, a naive baseline is pipelining the top-1 answer from one model to another, which performs worse than a single model.
%Instead of such selection, we leverage the top-$n$ QA pairs as additional knowledge, which will be described in the next section.
%In our work, we consider the retrieved top-$n$ QA pairs as additional context for Open-book QA.
%We 

To motivate this approach, Figure \ref{figure_preliminary} shows QA accuracy of our UR-QA and Recall of retrieved knowledge (recall@n) on the following variants of knowledge: 
(1) Document-only ($n$ passages); (2) Doc + QA history ($n+1$ passages).
Gains from adding one passage (concatenating $m=50$ QA history) suggest the complementary nature of QA history to documents, in terms of both QA and retrieval performances, regardless of the size
of retrieved passages $n$.

Inspired, we propose to combine
 $ \mathbf{d}_{1:n}$ and $\mathbf{k}$ as context, and a baseline \cite{oguz2020unik} concatenates all knowledge -- texts, tables, and knowledge graphs in the decoder.
Through this ``concat'' baseline, 
we can consider $\mathbf{k}$ of QA-pairs as (\text{n+1})th passage in Doc-QA, so that the final answer $A_{base}$ is obtained as follows:
\begin{equation}
    \begin{aligned}
    % \text{Fuse}(\mathbf{q}, \mathbf{d}_{1:n}, \mathbf{k}) = ~  \\
    & A_{base}(\mathbf{q}, \mathbf{d}_{1:n}, \mathbf{k}) = \\ 
    & ~~~~ \text{Dec}([\text{Enc}(\mathbf{q},\mathbf{d}_1);...;\text{Enc}(\mathbf{q},\mathbf{d}_n); \text{Enc}(\mathbf{q},\mathbf{k})])
    % \text{Dec}(\text{Enc}(\mathbf{q}, [\mathbf{d}_{1};...;\mathbf{d}_{n}; \mathbf{k}]))
    \end{aligned}
    \label{concat_baseline}
\end{equation}
where $[~; ]$ indicates the concatenation of encoder's outputs.
However, due to unreliable inputs from the concatenation, the performance may degrade with increasing noisy context, as reported in~\citet{oguz2020unik}. 
We hypothesize this as a cause of combining multi-knowledge underperforming a single model and propose selective QA.

\subsection{Selective UR-QA via Calibration}

Our distinction from concat baseline is that we compare the confidence of each answer from documents and QA history, then select the final answer $A_{ours}$ as follows: 
\begin{equation}
A_{ours} = \begin{cases}
  \mathbf{\hat{a}}_k & \text{if~} \text{Conf}(\mathbf{\hat{a}}_{k}|\mathbf{q},\mathbf{k}) \geq \text{Conf}(\mathbf{\hat{a}}_{d}|\mathbf{q},\mathbf{d}) \\
  \mathbf{\hat{a}}_d & \text{if~} \text{Conf}(\mathbf{\hat{a}}_k|\mathbf{q},\mathbf{k}) < \text{Conf}(\mathbf{\hat{a}}_d|\mathbf{q},\mathbf{d})
  \end{cases}
  \label{comparing_conf}
\end{equation}
where $\mathbf{\hat{a}}_k$ and $\mathbf{\hat{a}}_d$ are the decoded answers over QA pairs $\mathbf{k}$ and documents $\mathbf{d}$, respectively.
While the existing methods for confidence estimation adopt the likelihoods of language models,
to overcome its overfitting (Section 3.3.1), we propose two new measures, answerability (Section 3.3.2) and consistency (Section 3.3.3), 
to eventually ensemble these confidence estimates into a score.

\subsubsection{Sequence Likelihood of LM}

The key point of our method is to find the effective measurement of the answer confidence, which is essentially the calibration problem.
The confidence score $\text{P}(\mathbf{\hat{a}}|\cdot)$ should be able to discern the accurate answer, by comparing the reliability of each knowledge.
We propose the way to find such $\text{P}(\mathbf{\hat{a}}|\cdot)$ in the next paragraph, based on our analysis of the important factors on documents and QA-pairs.

Prior work \cite{hendrycks2016baseline} has proposed MaxProb -- a method that uses the maximum probability of a classifier as the confidence estimator for selective prediction.
For extractive QA, existing works \cite{zhang2021knowing,si2022revisiting} adopt MaxProb as a baseline, by using the sum of the maximum logits of the start and end of the answer span.
Meanwhile, we focus on calibrating generative language models, where its output is a token sequence.
To apply MaxProb for generative LMs, 
we select the maximum probability at each step by the argmax function in Eq. (\ref{eq3}), which can be viewed as greedy decoding.
The scores of decoded tokens are aggregate by product, as follows:

\begin{equation}
    \text{P}_{LM}(\mathbf{\hat{a}}|\mathbf{q},\mathbf{c}) =  \prod_{t=1}^{|\mathbf{\hat{a}}|} \text{P}_{LM}(\hat{a_t}|\mathbf{\hat{a}}_{<t}, \mathbf{q}, \mathbf{c})  % ^{1/|\mathbf{a}|}
    \label{lm_prob}
\end{equation}
where $\text{P}_{LM}(*)$ is the token probabilities obtained from LM head.
Since LM tends to underestimate the likelihood of longer texts, length normalization is essential as in \cite{adiwardana2020towards}.
To normalize as sequence lengths,\footnote{We empirically found that length normalization slightly improves the performance of Selective QA.} we take the geometric mean of the multiplicative terms, \ie, $\{ \text{P}_{LM}(\mathbf{\hat{a}}|\mathbf{q},\mathbf{c}) \}^{1/|\mathbf{\hat{a}}|}$.

However, this LM likelihood obtained by MaxProb has an inevitable problem.
MLE loss in Eq. (\ref{lm_loss}) enforces to train LM solely towards maximizing the likelihoods of observed sequences. 
Because the observed sequences (or labeled answers) can have diverse surface forms, MLE training inevitably leads to miscalibration.
In QA tasks, the sequence likelihood of QA models is reported to be often miscalibrated, or overconfident \cite{jiang2021can,kumar2019calibration}.

\begin{figure}[t]
	\centering
	\includegraphics[width=70mm]{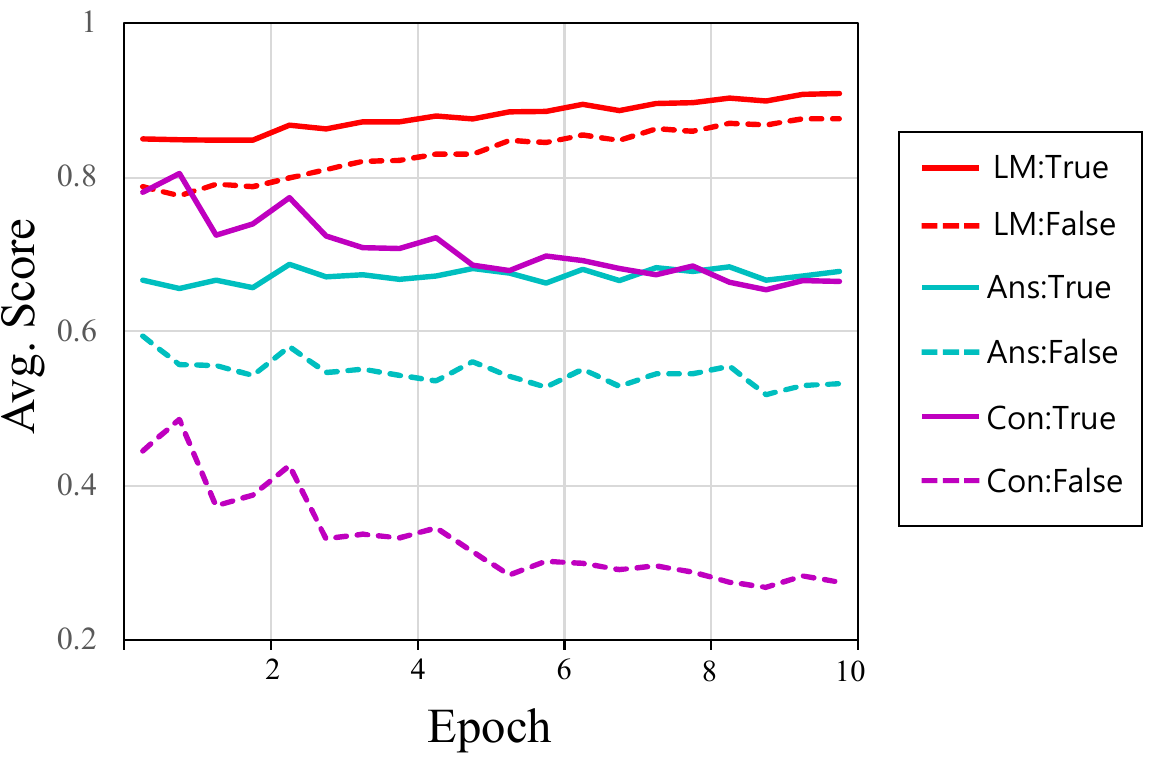}
	\caption{Average scores of three confidence estimates over training epochs. While the solid line is confidence on correct answers, the dashed line is confidence on incorrect answers. Red: LM likelihood, Aqua: Answerability, Purple: Consistency.}
	\label{calibation_overfitting}
\end{figure}

In Figure \ref{calibation_overfitting}, we also observe a consistent tendency in our open-domain QA task, where each line indicates the average confidence score of three estimates on correct predictions (solid line) and incorrect predictions (dashed line).
As the training steps increase, the scores of LM likelihood (red lines) increases monotonically, and even the gap between correct and incorrect predictions decreases.
We denote this problem as \textbf{calibration overfitting}, and hypothesize two causes (C1 and C2).
% \begin{itemize}
\begin{itemize}[leftmargin=0.4cm]
\setlength\itemsep{2pt}
\item C1: LM's objective maximizes the probabilities on answers regardless if the retrieved context is answerable or not, such that it is overconfident on unanswerable contexts.

\item C2: LM likelihood of a decoded output alone does not represent their uncertainty, while candidates unselected by greedy decoding can be a meaningful indicator of uncertainty.
\end{itemize}
 To deal with the above issues, we propose a new calibration approach of learning two measures: \textbf{Answerability} and \textbf{Consistency}, which are robust to calibration overfitting, as shown in Figure \ref{calibation_overfitting}.
 
\subsubsection{Answerability}
For \textbf{C1}, we learn an answerability score, ``P(Answerable)'', the probability that the passage can answer the given question, which has been studied in Machine Reading Comprehension tasks~\cite{rajpurkar2018know}.
% to retrieve a passage that can answer the question.
Our contribution is to train to predict answerability for the question/context pair $(\mathbf{q},\mathbf{c})$ 
for the purpose of detecting the low confidence when the given context $\mathbf{c}$ cannot answer question $\mathbf{q}$, \ie, unanswerable.
Training signals can be straightforwardly collected by whether $\mathbf{q}$ is answerable in $\mathbf{c}$, or not.
\begin{equation}
  \text{P(Answerable)} = \begin{cases}
  1, & \text{if~} \mathbf{q} ~ \text{is answerable in}~ \mathbf{c} \\
  0, & \text{otherwise~}
  \end{cases}
  \label{answerability}
\end{equation}

\subsubsection{Consistency} 

For \textbf{C2}, we learn a consistency score, ``P(Consistent)'', the probability of whether samples consistently match a correct answer.
The same decoded answer $\mathbf{\hat{a}}$ may have a high uncertainty, if a discarded candidate from the decoder is also highly plausible.
In contrast, the same answer has low uncertainty, if discarded candidates from the decoder are not plausible.

To estimate such sampling uncertainty, we apply sampling-based decoding (temperature=1) generating a set of samples of size $N$, and measure sampling consistency.
More formally, our supervision signal for uncertainty can be collected as:
\begin{equation}
    \text{P(Consistent)} = \frac{\sum_{i=1}^{N} \mathbbm{1}{(\mathbf{\hat{a}_i} = \mathbf{a}^*)}}{N}
\end{equation}
where $\mathbbm{1}()$ is 1 if the condition holds (0 otherwise). $\mathbf{\hat{a}}_i$ and $\mathbf{a}^*$ are $i$-th sampled output and the ground-truth, respectively.
$N$ is the number of samples, and we set $N=30$ in our experiment.

% To supervise the above scores, we can build individual regression estimators $\text{P}_f$ and $\text{P}_g$, so that,
% at inference time, we can use a calibration ensemble to combine the three scores:
% \begin{equation}
%     \begin{aligned}
%      \text{Confidence}(\mathbf{\hat{a}}) = ~ & \text{P}_{LM}(\mathbf{\hat{a}}|\mathbf{q},\mathbf{c}) +
%      \text{P}_f(\text{answerable}|\mathbf{x}) \\  + ~ &\text{P}_g(\text{consistent}|\mathbf{x})
%     \end{aligned}
% \end{equation}
% where $\mathbf{c}$ is the context of either document $\mathbf{d}$ or QA history $\mathbf{k}$.

\subsubsection{Prompted Calibration}

We then proceed to discuss the process of aggregating calibration components into a score, using LM for weak supervision.
%To supervise any scoring function, some works learn
LM has been used as a means of estimating scores by verbally expressing to estimate
a score as an output sequence, as adopted in diverse cases, \eg, sensibleness and safety \cite{thoppilan2022lamda} and uncertainty as question types \cite{lin2022teaching}.
The advantages of using a LM-based verbal estimator are twofold: (1) it eliminates the need to construct separate networks for scoring and (2) it captures the interdependency between answer prediction and its uncertainty within the same LM head.

To learn $S_\text{ans}$ and $S_\text{con}$ via verbal estimator, we convert the scores into discrete words.
Specifically, $S_{ans}$ is expressed as either \textit{True} or \textit{False}. 
The continuous values $S_{con}$ in training data are sorted and partitioned into equally sized quantiles (\ie, \textit{High}, \textit{Medium}, and \textit{Low}).
Then, we train \ourmodel{} to generate the output template, prompted with the verbalized scores, as follows:
% Specifically, we express $y_{ans}$ as either \textit{True} or \textit{False}, and convert real value $y_{con}$ into discrete words -- by partitioning it into equally sized quantiles (\ie, \textit{High}, \textit{Medium}, and \textit{Low}), as the following prompt:
\begin{figure}[h]
	\centering
	\includegraphics[width=72mm]{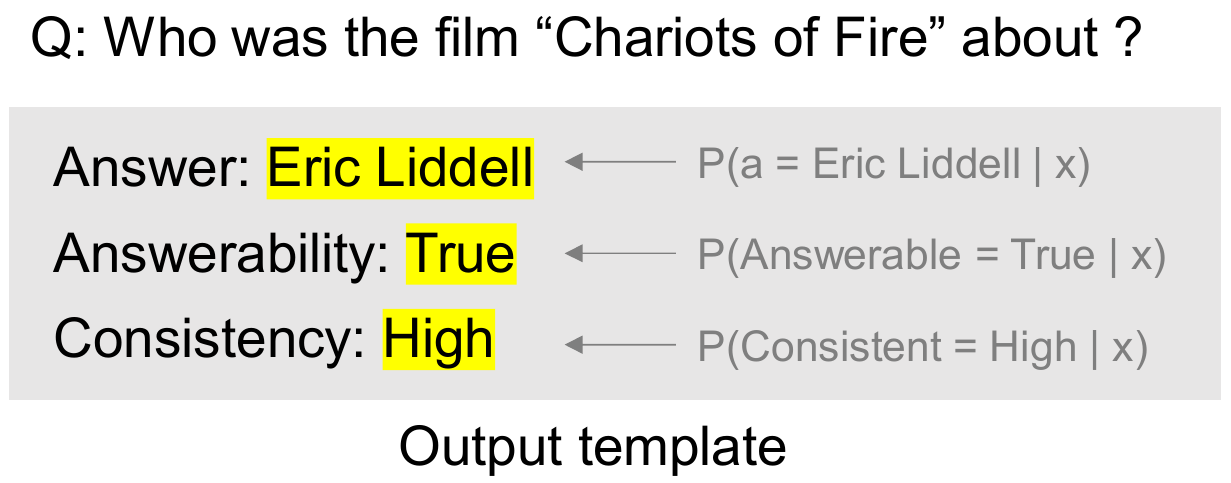}
	\label{figure2}
\end{figure}

After training with the prompt, we can estimate $S_\text{ans}$ and $S_\text{con}$ on test examples, through the likelihood of token ``True'' or ``High'', as follows:
% two likelihoods of $\text{P}_f$ and $\text{P}_g$ at the inference time, as follows: 
\begin{equation}
    \begin{gathered}
    \text{P(Answerable)} = \text{P}_{LM}(\text{True}| \mathbf{y}_{<True},\mathbf{q},\mathbf{c}) \\
    \text{P(Consistent)} = 1 \cdot \text{P}_{LM}(\text{High}| \mathbf{y}_{<High},\mathbf{q},\mathbf{c}) \\ + 0.5 \cdot \text{P}_{LM}(\text{Medium}| \mathbf{y}_{<Medium},\mathbf{q},\mathbf{c}) \\
    \end{gathered}
\end{equation}
where $\mathbf{x}$ is the context with the above prompt.
At inference time, we can use a calibration ensemble by averaging the three scores:
\begin{equation}
    \begin{gathered}
     \text{Conf}(\mathbf{\hat{a}}|\mathbf{q},\mathbf{c}) = ~ \frac{1}{3} 	\Big( \text{P}_{LM}(\mathbf{\hat{a}}|\mathbf{q},\mathbf{c}) \\ ~~~ + \text{P(Answerable)}  + ~ \text{P(Consistent)} \Big)
    \end{gathered}
    \label{eq:ensemble}
\end{equation}
This final confidence is used in Eq. (\ref{comparing_conf}) for comparing two candidates, then it decides the final answer.
\section{Experiment}
In our experiments, we first demonstrate that our proposed confidence scores effectively improve the calibration for question answering. We then examine how these scores contribute to an overall improvement in question answering performance. Finally, we provide qualitative analysis to gain a deeper understanding and insight on our method.

\paragraph{Datasets}
% We use the open-domain QA version of Natural Questions (NQ) \cite{kwiatkowski2019natural} and TriviaQA (TQA) \cite{joshi2017triviaqa} following the previous setting ~\cite{karpukhin2020dense,izacard2021leveraging}.\footnote{https://github.com/facebookresearch/FiD}
% For a collection of knowledge, we use PAQ database for QA pairs \cite{lewis2021paq}, and Wikipedia for documents \cite{karpukhin2020dense}.
We use the open-domain QA version of 
Natural Questions \cite{kwiatkowski2019natural} and TriviaQA  \cite{joshi2017triviaqa}, following the previous setting ~\cite{karpukhin2020dense,izacard2021leveraging}.\footnote{https://github.com/facebookresearch/FiD}
The details of the benchmarks are as follows:

\begin{itemize}[leftmargin=0.4cm]
\setlength\itemsep{2pt}
    \item \textbf{Natural Questions (NQ)} contains real user questions from Google search engine. We use training/dev/testing splits for open-domain question answering, consisting of 79K train, 8.7k dev, 3.6K test examples.
    \item \textbf{TriviaQA (TQA)} is constructed from web-scraped trivia questions. We use TriviaQA open-domain training/dev/testing splits, consisting of 79K train, 8.8k dev, and 11K test examples.
\end{itemize}

\paragraph{Implementation}
We implement our models upon T5 with the size of 770M (or `Large') and 3B 3B (or `XL'), and fine-tune them on NQ and TQA.
To retrieve the contexts ($\mathbf{d}$ and $\mathbf{k}$), we use the same off-the-shelf retrieval as used by baselines: FiD-KD \cite{izacard2020distilling} for DR, and RePAQ \cite{lewis2021paq} for QR.
While FiD-KD set the number of passages to 100, we used top-50 passages for DR-QA due to GPU limitations, which is the reason why our DR-QA performed lower than FiD-KD. For QA-history, we concatenate top-50 QA-pairs into a single passage.
We use 8 Tesla A100 40GB GPUs for all experiments.

To retrieve the contexts ($\mathbf{d}$ and $\mathbf{k}$), we use the same off-the-shelf retrieval as used by baselines: FiD-KD \cite{izacard2020distilling} for Doc-QA, and RePAQ \cite{lewis2021paq} for QR.
For a collection of knowledge, we also use PAQ database for QA pairs \cite{lewis2021paq}, and Wikipedia for documents \cite{karpukhin2020dense}.
Table~\ref{retrieval} shows the accuracy of retrievals from documents and QA-pairs.
If a correct answer is included in the top-\textit{K} contexts, the retrieval is assumed to succeed.
While this measure calculated by naive string matching is commonly used in \cite{karpukhin2020dense,izacard2021leveraging,izacard2020distilling}, it is not perfect as false negative examples can be counted as true positive.

\begin{table}[t]
\centering
\setlength{\tabcolsep}{6pt}
\scalebox{0.94}{
\begin{tabular}{c|cc|cc}
\noalign{\hrule height 1pt} 
\multicolumn{1}{c|}{\multirow{2}{*}{\textbf{Metric}}} & \multicolumn{2}{c|}{\textbf{Documents}} & \multicolumn{2}{c}{\textbf{QA-Pairs}}  \\
\multicolumn{1}{c|}{} & NQ & TQA  & NQ & TQA 
 \\
% \multicolumn{1}{c}{Metric}  & NQ   & TQA  & NQ   & TQA   \\ 
\noalign{\hrule height 1pt}

Top-1 & 50.9 & 56.9  & 41.7  &  41.3  \\
Top-5  & 75.1 & 80.2 & 53.5 &  51.2    \\ 
Top-10 & 80.8 & 84.8 & 58.5 & 55.7 \\  
Top-30 & 86.8 & 88.6 & 64.5 & 61.4   \\  
Top-50 & 88.7 & 89.7 & 67.2 & 64.0  \\

% Top-1 & 50.91 & 56.86  & 41.83  &  41.81   \\
% Top-5  & 75.12 & 80.22 & 53.46 &  51.15    \\ 
% Top-10 & 80.80 & 84.77 & 58.45 & 55.71  \\  
% Top-30 & 86.81 & 88.59 & 64.49 & 61.43     \\  
% Top-50 & 88.67 & 89.66 & 67.15 & 63.98   \\
\noalign{\hrule height 1pt} 
\end{tabular}}
\caption{Retrieval accuracy on test sets in NQ and TQA.}
\label{retrieval}
\end{table}

% For development set, we used provided set~\cite{}, and select the maximum harmonic mean of their accuracies as a best model

\paragraph{Baselines}
To show the effectiveness of our method, 
we compare previous models over a single source -- FiD \cite{izacard2021leveraging}, FiD-KD \cite{izacard2020distilling}, UnitedQA \cite{cheng2021unitedqa}, and R2-D2 \cite{fajcik2021r2} over documents, and RePAQ \cite{lewis2021paq} over QA-pairs.
``Our backbone'' is reimplemented from FiD-KD, while the difference is the number of retrieved documents.
To validate the complementary of documents and QA-history, we compare UR-QA on a single source without our selection: ``Document Only'' and ``QA-History Only''.
% As the backbone models, we denote QA over retrieved documents as \textbf{DR-QA} (reimplemented FiD-KD)
% and OBQA over retrieved QA pairs as \textbf{QR-QA}.
% is FiD model over QA pairs retrieved from Question Retrieval.
% we observe the improvement over (1) QA pair-only baselines (RePAQ variants) and (2) passage-only baselines. 
As baselines over multiple sources, we compare our method with ``Base1: Pipeline'' consisting of RePAQ and FiD \cite{lewis2021paq}, and ``Base2: Concat'' in Eq. (\ref{concat_baseline}), inspired by \cite{oguz2020unik}.

\begin{table}[t]
\centering
\scalebox{0.88}{
\begin{tabular}{l|cc}
\noalign{\hrule height 1pt} 
\multicolumn{1}{c}{\textbf{Method}}  & \textbf{NQ}   & \textbf{TQA}   \\ 
\noalign{\hrule height 1pt}
\textit{Document-based QA} & & \\
~~ RAG & 44.5 & 56.8  \\
~~ UnitedQA & 54.7 & 70.5  \\
~~ R2-D2  & 55.9 & 69.9  \\
~~ FiD ($n$=100, Large) & 51.4 & 67.6  \\
~~ FiD-KD ($n$=100, Large) & 54.4 & 72.5  \\
~~ Our backbone ($n$=50, Large) & 53.4 & 71.4  \\
% & ours                             &      &       \\ 
\hline
\textit{QA as Retrieval} & & \\
~~ TF-IDF & 22.2 & 23.5  \\
~~ RePAQ (Retriever only) & 41.7 & 41.3  \\
~~ RePAQ (Reranker) & 47.6 & 52.1  \\
\hline
\textit{UR-QA (on a single source)} & & \\
~~ Document Only ($n$=10, Large)  & 50.7 & 69.2  \\
~~ Document Only ($n$=50, Large)  & 53.5 & 71.3  \\
~~ Document Only ($n$=50, XL)  & 56.0 & 73.5  \\
~~ QA-History Only (Large)  & 46.6 & 54.3  \\
~~ QA-History Only (XL)  & 47.7 & 56.8  \\
\hline
\textit{UR-QA (Document + QA-History)} & & \\
~~ Base1: Pipeline (Large)  & 52.3 & 67.3  \\
~~ Base2: Concat (Large)  & 53.9 & 72.0      \\
~~ Base2: Concat (XL)  & 56.7 & 74.2     \\
%~~ Ours on DR-QR (Sep, Large)  & 56.0   &  72.8      \\
~~ Ours: SelectiveQA ($n$=10+1, Large)  & 53.6 & 70.6     \\
~~ Ours: SelectiveQA ($n$=50+1, Large)  & 55.4 & 72.6     \\
%~~ Ours on DR-QR (Sep, Xlarge)  & 58.1 & \textbf{74.7}     \\
~~ Ours: SelectiveQA ($n$=50+1, XL)    & \textbf{58.2} &  \textbf{74.5}    \\ \hline
% ~~ Oracle - Upper Bound (large) & 62.7 & 75.4    \\
% ~~ Oracle - Upper Bound (3B) &  &     \\
\noalign{\hrule height 1pt} 
\end{tabular}}
\caption{Comparison to open-domain QA models on NQ and TQA. Note that while FiD and FiD-KD use 100 documents, we use 10 or 50 documents for ours.}
%We report the result of our `separate (Sep)' models where we use individual LMs for each knowledge source, together with `single' models where both documents and QA pairs are handled by one LM.}
\label{main_result}
\end{table}

\paragraph{Main results}
Table \ref{main_result} shows the performance of our models, with comparable other models in NQ and TQA.
We evaluate the performance of our models by Exact Match (EM) score, which is a standard metric for open domain question answering \cite{izacard2021leveraging}.
Our models outperform the baseline models for both datasets and in both model sizes (Large and XL readers).
In NQ, we observe that our selective UR-QA achieved the performance gain of $1.9$ EM over UR-QA (``Document Only''), and $8.8$ over UR-QA (``QA-History Only''), on T5-Large.
% the performance gain of $1.9$ EM over DR-QA (Large), and $8.8$ over QR-QA (Large) built on T5-large.
Our method (Large-NQ) also outperforms Base1: Pipeline \cite{lewis2021paq} by $2.9$ and Base2: Concat by $1.5$, respectively.
Our best model with larger size (XL) shows \textbf{58.2} EM in NQ, which is the highest among the compared models.
Meanwhile, our model trained on TQA (Large-TQA) increases EM score by $0.9$ over UR-QA (``Document Only'') baseline, and $17.9$ over UR-QA (``QA-History Only'').
Our best performing model in TriviaQA (XL-TQA) achieves the highest score as well, recording \textbf{74.5} EM.
%We report the result of our `separate (Sep)' models where we use individual LMs for each knowledge source, together with `single' models where both documents and QA pairs are handled by one LM.}

% Alternatively, when we can afford to keep two separate models for two knowledge sources, results are comparable or slightly better: \textbf{56.0} (Large-NQ), 58.1 (XL-NQ), \textbf{72.8} (Large-TQA), and \textbf{74.7} (XL-TQA),
% where higher scores than those reported from a single model in Table \ref{main_result} are shown in \textbf{bold}.
% We stress that a single model, incurring a lower cost, is better for NQ and  only marginally lower by 0.6, 0.6, and 0.2, respectively for (Large-NQ), (Large-TQA), and (XL-TQA), and report the results from separate models from this point on.

\begin{table}[t]
\setlength{\tabcolsep}{3pt}
\centering
\scalebox{0.9}
{
\begin{tabular}{l|cc|cc}
    \noalign{\hrule height 1pt}
    \multicolumn{1}{c|}{\multirow{2}{*}{\textbf{Method}}} & \multicolumn{2}{c|}{\textbf{NQ}} & \multicolumn{2}{c}{\textbf{TQA}} \\
    \multicolumn{1}{c|}{} & ECE{\textit{$_\downarrow$}}  & AUC{\textit{$_\downarrow$}}  & ECE{\textit{$_\downarrow$}} & AUC{\textit{$_\downarrow$}} \\
    \noalign{\hrule height 1pt}
    FiD-KD (LM likeli) & 0.310 & 0.251 & 0.186 & 0.103 \\ 
    ~ +Temp Scaling & 0.246 & 0.247 & 0.063 & 0.098\\ 
    \hline
    \textsc{UR (Doc-only)} & & & &  \\ 
    ~ (1) LM likelihood & 0.305  & 0.290  & 0.182 & 0.091  \\ 
    ~ (2) Answerability & 0.154   & 0.307 & 0.185 & 0.116 \\ 
    ~ (3) Consistency & \textbf{0.134} & 0.244  & \textbf{0.154} & 0.099 \\ 
    ~ (1+2+3) Ours & 0.163  & \textbf{0.240} & 0.168 & \textbf{0.088} \\ 
    \hline
    \textsc{UR (QA-only)} & & & &  \\ 
    ~ (1) LM likelihood & 0.396 & 0.390  & 0.326 & 0.209  \\ 
    ~ (2) Answerability & \textbf{0.126}  & 0.293 & 0.174 & 0.188  \\ 
    ~ (3) Consistency & 0.153 & 0.298 & \textbf{0.074} & 0.174 \\ 
    ~ (1+2+3) Ours & 0.147 & \textbf{0.289} & 0.170 & \textbf{0.171}  \\ 
    \noalign{\hrule height 1pt}
\end{tabular}
}
\caption{Calibration Evaluation: ECE \& AUC of our methods, compared to FiD-KD. ${\downarrow}$ means the lower the metric, the better the calibration is.}
\label{tab:calibration}
\end{table}

% \input{tables/calibration2.tex}

%Table 2 shows the effect of our calibration method in Natural Questions and TriviaQA.
%For the evaluation, we use two metrics for calibration performance: Expected Calibration Error (ECE) and Area Under Curve (AUC) of the risk-coverage plot.
%The former indicates how much the confidence score deviates from the accuracy in average \cite{minderer2021revisiting}, while the latter indicates how much the confidence score retain the correct answers in SelectiveQA setting \cite{el2010foundations}.
%Note that both metrics indicate better calibration when lower.
%(See Appendix A for more detailed definition of each metric)

%which indicates how much the expected accuracy deviates from the expected confidence score (cite),
%where both indicates better calibration when lower.
%(More detailed explanation on metrics will be provided below)

\paragraph{Does our method improve calibration for open-domain QA?}
We use two metrics for the evaluation of the calibration performance: Expected Calibration Error (ECE) and Area Under Curve (AUC) of the risk-coverage graph.
ECE is one of the most commonly used metric in previous works \cite{guo2017calibration,minderer2021revisiting,si2022revisiting,jiang2021can}, which indicates how much the expected accuracy deviates from the expected confidence score.
We use the density-based ECE from \citet{minderer2021revisiting}, defined as below:
\begin{align}
    \mathrm{ECE}&=\sum^{M}_{m=1} \frac{1}{M} |\mathrm{Acc}(B_m) - \mathrm{Conf}(B_m)|,
\end{align}
where $M$ is the total number of bins (we use $M=10$), $B_{m}$ denotes $m$-th bucket, Acc($B_{m}$) is the mean accuracy of $B_{m}$, and Conf($B_{m}$) is the mean confidence.
In density-based ECE, an equal number of predictions are assigned to each bin.

On the other hand, the risk-coverage plot \cite{wang2017safer} shows the trade-off between the coverage and risk, where the former is measured as the fraction of test cases that model makes prediction on, and the latter is the error rate (or 1$-$accuracy) at that coverage.
Specifically, the risk is reportedly high when the coverage increases \cite{el2010foundations}, since the less confident examples come into consideration.
Lower AUC of risk-coverage plot indicates the lower average risk, which means more chance of retaining correct answers in selectiveQA.

\begin{figure}[t]
	\centering
	\includegraphics[width=65mm]{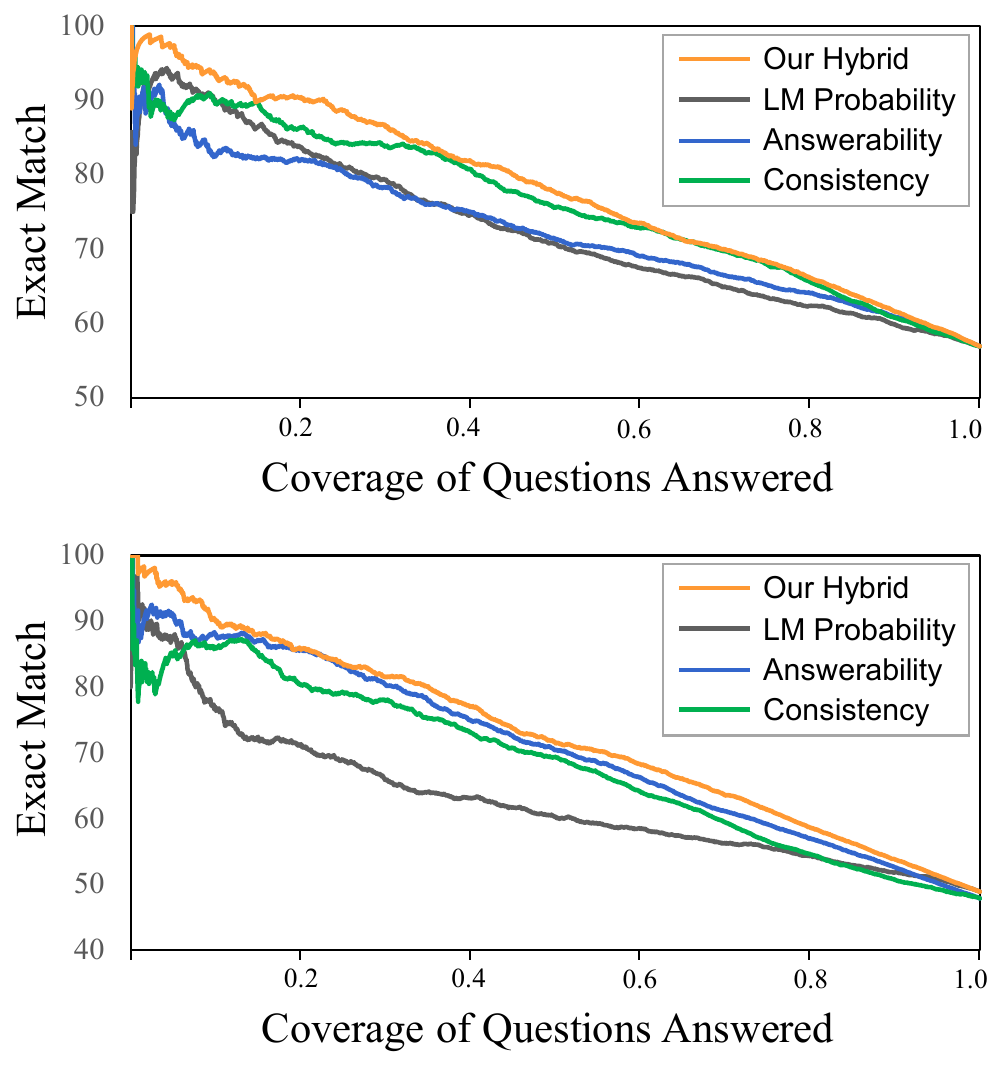}
	\caption{Accuracy-coverage plot (NQ Large). Ours retains the highest accuracy for all coverage. Top: UR-QA (Doc Only), Bottom: UR-QA (QA-History Only).}
    % \vspace{-2mm}
	\label{figure3}
\end{figure}

Table \ref{tab:calibration} shows that our method (1+2+3) has the lowest AUC in all observed cases.
Ours robustly outperforms individual measures in AUC, while there is no `all-time winner' among individual measures.
The robustness of our method is observed in ECE as well -- ours is the second-lowest in all cases, while the ranking of others shifts with the change of the dataset or knowledge source.
Meanwhile, we attempted temperature scaling \cite{guo2017calibration} by optimizing a scaling factor in [0,10], but observed no significant improvement on AUC.

Figure \ref{figure3} provide a finer-grained illustration of this situation,
where our hybrid (1+2+3) has the best accuracy (Exact Match) for all coverage in both documents and QA pairs, while the accuracy of other measures fluctuates beneath it.

\paragraph{Does better calibration improve the complementarity of two knowledge sources?}
Our goal is to enhance the complementarity of documents and QA-history through better calibration, leading to improved QA performance. We investigate if improved calibration truly contributes to the utilization of complementarity. 
As seen in Table \ref{ablation}, our hybrid (1+2+3) method, which exhibits the best calibration performance in Figure \ref{figure3}, proves to be the most effective criterion for selection, while language model likelihood often fails to improve QA performance beyond the baseline.
To examine the upper bound of our approach, we also report the ideal QA performance (`Oracle') which is attainable with the perfect selection.
The results indicate that there is a significant potential for complementarity to further enhance QA performance, and that the selection method plays a crucial role in realizing this potential gain.

\begin{table}[t]
\centering
\scalebox{0.9}{
\begin{tabular}{c|l|cc}
\noalign{\hrule height 1pt} 
Size & \multicolumn{1}{c|}{Method}  & NQ   & TQA   \\ 
\noalign{\hrule height 1pt}

\multirow{6}{*}{Large} & Ours: (1) LM likelihood  & 52.2  & 70.4   \\
& ~~~~~~~~~ (1) + Temp Scaling   & 52.1   &  70.5  \\
& Ours: (2) Answerability    & 55.1 &  71.7    \\ 
& Ours: (3) Consistency    & 54.9  & 72.4  \\  
& Ours: (1+2+3)     & \textbf{56.0} & \textbf{72.8}     \\  
& Oracle - Upper Bound & 62.7 & 75.5    \\
\hline
\multirow{6}{*}{Xlarge} & Ours: (1) LM likelihood     & 54.5  & 73.5      \\
& ~~~~~~~~~ (1) + Temp Scaling & 54.5 & 73.6     \\
& Ours: (2) Answerability  & 57.6 & 74.2     \\ 
& Ours: (3) Consistency    & 57.0 & 74.3     \\  
& Ours: (1+2+3)    & \textbf{58.1} & \textbf{74.7}     \\  
& Oracle - Upper Bound  & 64.6 & 77.5    \\
\noalign{\hrule height 1pt} 
\end{tabular}}
\caption{Ablation Study}
\label{ablation}
\end{table}

\paragraph{Is ours robust under domain shifts?}
To ensure that our model is robust under domain shifts, we conducted cross-evaluation by out-of-domain evaluations: evaluating our QA model (trained on the NQ dataset) on the TQA test set and our QA model (trained on the TQA dataset) on the NQ test set. As shown in Table \ref{domain_shift}, we found that utilizing both knowledge sources is more beneficial than using a single source, even under domain shifts. Our proposed selective UR  achieved gains of 3.8 EM on the NQ dataset and 2.1 EM on the TQA dataset, compared to baselines that used a single source.

\begin{table}[h]
\centering
\scalebox{0.90}{
\begin{tabular}{c|cc}
\noalign{\hrule height 1pt} 
Method & \makecell{Train on TQA \\ Eval on NQ}   & \makecell{Train on NQ \\ Eval on TQA}	   \\ 
\noalign{\hrule height 1pt}
UR (Doc-only) & 34.1 & 59.9   \\
UR (QA-only) & 35.2 & 49.0  \\ 
Selective UR & 39.0 & 62.0 \\  

\noalign{\hrule height 1pt} 
\end{tabular}}
\caption{Results under domain shift}
\label{domain_shift}
\end{table}

\begin{figure}[t]
	\centering
	\begin{subfigure}[t]{\linewidth}
	    \centering
		\includegraphics[width=75mm]{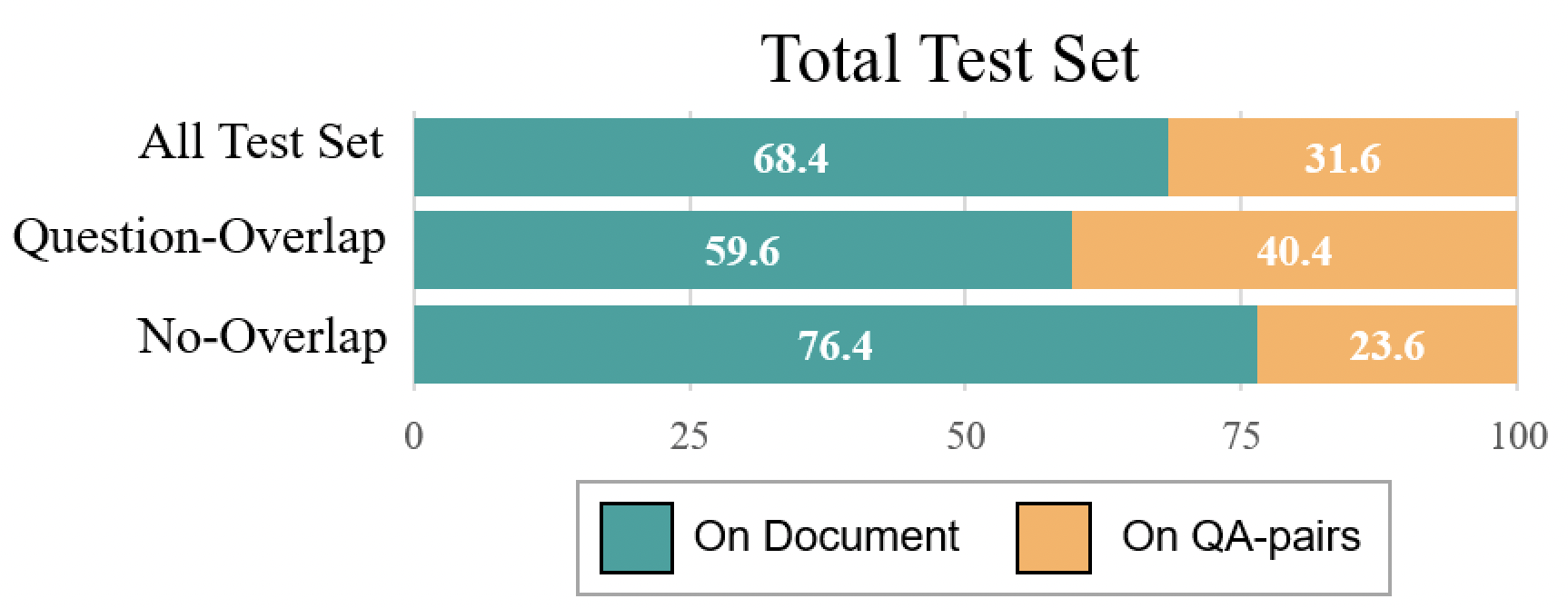}
		\caption{The results on all test set}
	\end{subfigure}
	\begin{subfigure}[t]{\linewidth}
		\includegraphics[width=75mm]{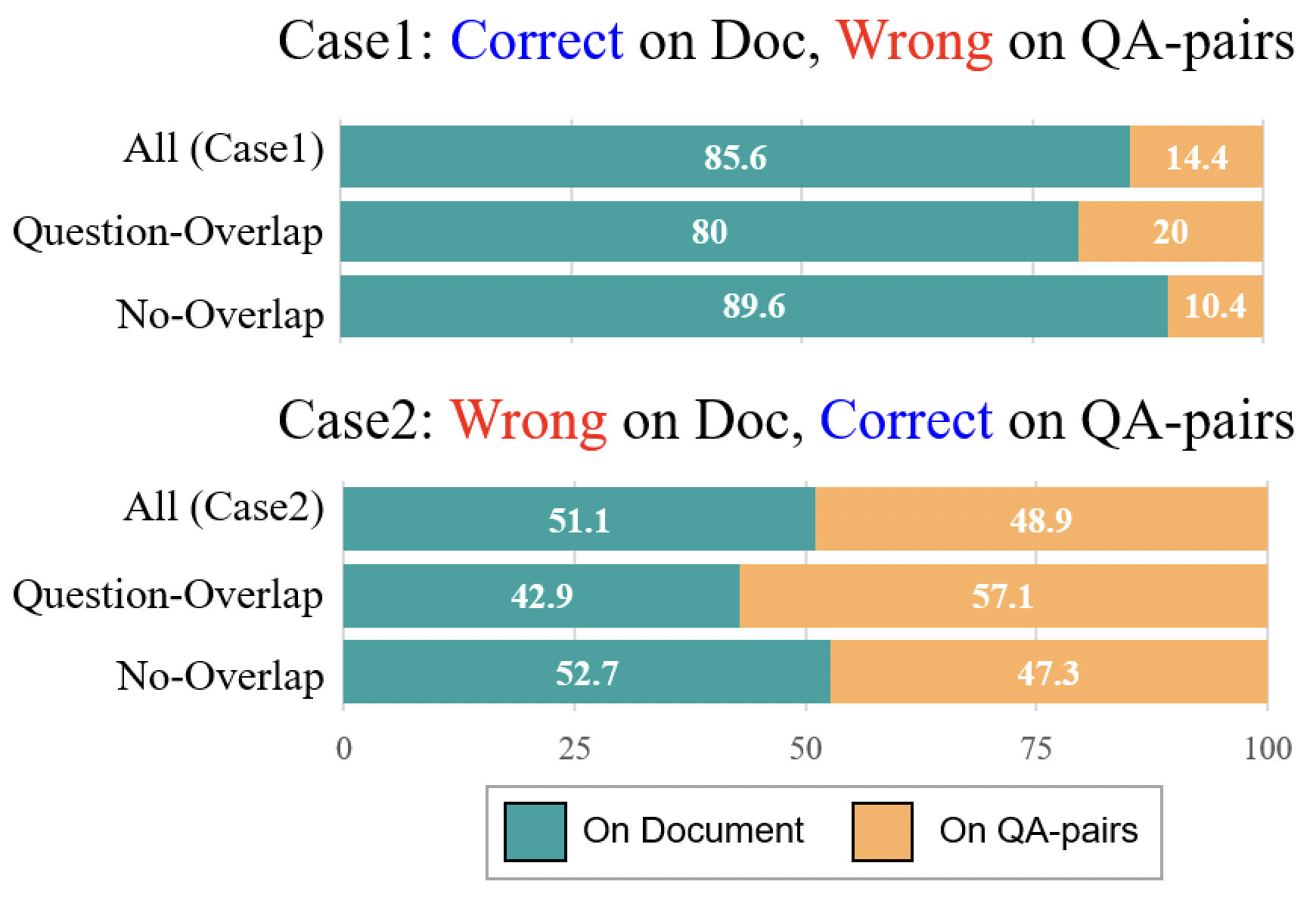}
		\caption{The results on critical cases 1$\&$2}
	\end{subfigure}
	\caption{Selection ratio of each knowledge source, from the result of NQ large model.}
	\label{selection_ratio}
\end{figure}

\paragraph{Model's Selection Ratio}
We remark our model's behavior that is related to the generalization.
Previous work~\cite{lewis2021question} splits test set into  paraphrased questions in training set (``Question-Overlap''), and unseen questions (``No-overlap'').
On the divided sub sets, we observe which knowledge (either documents or QA pairs) our method selected.
% how effectively models match a target question with paraphrases, or generalize on unseen questions.
% Figure \ref{selection_ratio} shows results on test set 
% splits that measure how effectively models match a target question with existing QA-pairs in train set (``Q overlap''), and generalize to unseen questions (``No overlap'').
Figure \ref{selection_ratio} shows the selection ratio of on total test set and Question-overlap/No-overlap sets.
As shown in Figure \ref{selection_ratio} (a), our method tends to select document knowledge (68.4\% on all test set).
On the question-overlap set, the ratio of selecting QA-pair knowledge increased on the Question-overlap set (31.6\% $\rightarrow$ 40.4\%).
This means the tendency of selecting QA pair knowledge increased when knowledge matching with questions in training set.
In contrast, on the no-overlap set, the tendency of selecting documents increased (68.4\% $\rightarrow$ 76.4\%), which means reading documents is more preferred for generalization on unseen questions.

For a closer look, we select only \textit{critical cases} where only one of the candidate answers is correct -- Case1: the answer from documents is correct, but one from QA-history is wrong, and Case2: one from documents is wrong, but one from QA-history is correct.
As shown in Figure \ref{selection_ratio}(b), in Case1, document is the majority of the selection, which increases the complementarity of the two knowledge.
Meanwhile, in Case2, the ratio of selecting documents (51.1\% on all Case2) is the error rate, which is potential room for improvement in our selection.
% use more QA pairs for the paraphrased question of train example, and more documents for truly unseen questions.

% This trend is consistently observed in critical cases for selection, where only one of the candidate answers is correct. 
% When the given questions are the paraphrase of train examples, 
% the EM score rises from $48.9$ to $57.1$ in the test set split where the correct answers come from QA pairs only.
% This observation gives an insight for handling the well-known trade-off between memorization and generalization: 
% when the model's ability of memorization and generalization differs over multiple data sources, we can leverage the difference to overcome the limits of individual sources, with an aid of careful selection.

\section{Acknowledgement}
This work was supported by LG AI Research. 	
This work was partly supported by Institute of Information \& communications Technology Planning \& Evaluation (IITP) grant funded by the Korea government(MSIT) [NO.2021-0-01343, Artificial Intelligence Graduate School Program (Seoul National University)].

\section{Conclusion}

This paper studies the selective QA system leveraging both document and QA-pair corpus.
For careful selection, we propose a novel and effective calibration method based on Answerability and Sampling Consistency and leverage them for comparing and selecting two knowledge sources.
On two benchmarks: NQ and TQA, we empirically show our proposed methods outperform existing approaches for open-domain question answering tasks.

\section{Limitations}
We have identified several limitations in our work and propose future directions to improve them:

(i) The  sources for UR-QA in this paper are limited to the document corpus and QA-history, but our unified reader is not restricted to take specific sources. Further research can explore the generalizability of UR-QA to more diverse sources, such as linearized knowledge sources as proposed in \cite{oguz-etal-2022-unik}. 
Future work can also explore the optimal method for considering LM likelihood, answerability, and consistency together. 
% (ii) We use simple averaging to compute the confidence ensemble in Eq.  \ref{eq:ensemble}, where there can exist multiple ways to integrate the scores. 

(ii) Though it is not the focus of this work to optimize readers,  our proposed UR-QA can orthogonally benefit from improvement in retrieval. Further study on the retrieval for UR-QA can be conducted, including the direction to co-optimize the reader and retriever as proposed in \cite{izacard2020distilling}.

\bibliography{anthology,custom}
\bibliographystyle{acl_natbib}

% \clearpage
% \input{7_appendix}

\end{document}